\documentclass[conference,letterpaper]{IEEEtran}
\IEEEoverridecommandlockouts
\usepackage{fancyhdr}
\usepackage[dvips]{graphicx}
\usepackage{indent}
\usepackage{amsmath,amssymb}
\usepackage{paralist}
\usepackage{eclbkbox}
\usepackage{subfigure}
\renewcommand{\baselinestretch}{0.9} 
\makeatletter
\def\tbcaption{\def\@captype{table}\caption}
\def\figcaption{\def\@captype{figure}\caption}
\makeatother

{\pointlessenum\begin{enumerate}}%
{\end{enumerate}}

\def\borderarray#1#2#3#4#5#6{%
\setbox0\hbox{$\begin{array}{#5}#6\end{array}$}
\setlength{\dimen1}{\wd0}\addtolength{\dimen1}{-#3}\addtolength{\dimen1}{-\arraycolsep}
\setlength{\dimen2}{\ht0}\addtolength{\dimen2}{-#4}
\setbox1\hbox{$\left#1\rule{\dimen1}{0pt}\rule{0pt}{\dimen2}\right#2$}
\setbox0\hbox{\raisebox{\dp0}{\box0}\kern-\dimen1\kern-5pt\raisebox{\dp1}{\box1}}
\vcenter{\box0}
}

\usepackage{fancyhdr}

\begin{document}
\title{A Proposal of Interactive Growing Hierarchical SOM
\thanks{\copyright 2011 IEEE. Personal use of this material is permitted. Permission from IEEE must be obtained for all other uses, in any current or future media, including reprinting/republishing this material for advertising or promotional purposes, creating new collective works, for resale or redistribution to servers or lists, or reuse of any copyrighted component of this work in other works.}
}

\author{\IEEEauthorblockN{Takumi Ichimura}
\IEEEauthorblockA{Faculty of Management and Information Systems,\\
Prefectural University of Hiroshima\\
1-1-71, Ujina-Higashi, Minami-ku,\\
Hiroshima, 734-8559, Japan\\
Email: ichimura@pu-hiroshima.ac.jp}
\and
\IEEEauthorblockN{Takashi Yamaguchi}
\IEEEauthorblockA{Department of Information Systems,\\
Tokyo University of Information Sciences\\
4-1 Onaridai, Wakaba-ku, Chiba, 265-8501 Japan\\
Email: tyamagu@rsch.tuis.ac.jp}
}

\maketitle

\fancypagestyle{plain}{
\fancyhf{}	
\fancyfoot[L]{}
\fancyfoot[C]{}
\fancyfoot[R]{}
\renewcommand{\headrulewidth}{0pt}
\renewcommand{\footrulewidth}{0pt}
}

\pagestyle{fancy}{
\fancyhf{}
\fancyfoot[R]{}}
\renewcommand{\headrulewidth}{0pt}
\renewcommand{\footrulewidth}{0pt}

\begin{abstract}
Self Organizing Map is trained using unsupervised learning to produce a two-dimensional discretized representation of input space of the training cases. Growing Hierarchical SOM is an architecture which grows both in a hierarchical way representing the structure of data distribution and in a horizontal way representation the size of each individual maps. The control method of the growing degree of GHSOM by pruning off the redundant branch of hierarchy in SOM is proposed in this paper. Moreover, the interface tool for the proposed method called interactive GHSOM is developed. We discuss the computation results of Iris data by using the developed tool.
\end{abstract}


\IEEEpeerreviewmaketitle

\section{Introduction}
The current information technology can collect various data sets  because the recent tremendous technical advances in processing power, storage capacity and network connected cloud computing. The sample record in such data set includes not only numerical values but also language, evaluation, binary data such as pictures. The technical method to discover knowledge in such databases is known to be a field of data mining and developed in various research fields.

The data mining is seen as an increasingly important tool by modern business to transform unprecedented quantities into business intelligence giving an informational advantage. Some data mining tools are currently used in a wide range of profiling practices such as marketing, fraud detection and medical information. The traditional method in data mining tools includes Bayes' theorem and regression analysis. As data sets have growth in size and complexity, automatic data processing technique in the field computer sciences is required. The dataminig in the field of neural networks(NNs), clustering, genetic algorithms, decision trees and support vector machines come into existence in the research of innovative soft computing methodologies.

Data mining is the process of applying these methods to data with the intention of uncovering hidden patterns. An unavoidable fact of data mining is that the subsets of data being analyzed may not be representative of the whole domain, and therefore may not contain examples of certain critical relationships and behaviors that exist across other parts of the domain. Moreover, the probability of inclusion of missing data and/or contradictory data becomes high because the means or instrumentality for storing information is to store the raw data in the storage by the automated collecting process.

Self organizing map (SOM)\cite{Kohonen95} is a type of artificial neural network that is trained using unsupervised learning to produce a low dimensional, discretized representation of the input space of the training samples, called a map. SOM is known to be an effective clustering method because it can learn regardless of data size and can intuitively show clustering results visually using maps. However, the clustering result by SOM has ambiguity because the boundary of clusters is not clear. In order to improve the clustering capability, \cite{Yamaguchi08a,Yamaguchi08b,Yamaguchi11} proposed the adaptive tree structured clustering (ATSC) in order to clarify clustering result of SOM. ATSC is divisive hierarchical clustering algorithm (DHCA) that recursively divides a data set into 2 subsets using SOM. In each partition, the data set is quantized by SOM and the quantized data is divided using agglomerative hierarchical clustering algorithm (AHCA). \cite{Yamaguchi08a,Yamaguchi08b} reported that ATSC can extract a tree structure that include potential hierarchical relationship without decreasing SOM classification performance within feasible time. Moreover, \cite{Yamaguchi10,Ichimura11} reported the search method of the optimal number and its layout in each map simultaneously.

Rauber et al. proposed an basic algorithm of the growing hierarchical self organizing map (GHSOM)\cite{Rauber02}. The algorithm has been chosen for this application for its capability to develop a hierarchical structure of clustering and for the intuitive outputs which help the interpretation of the clusters. However, GHSOM divides a data set into sub clusters immoderately if the distribution of samples is complex. There is a trade off for human designers between the investigation of the shape of a partial detailed cluster and the entire distribution of samples. In order to grasp an overview of tree structure of GHSOM, we propose an interactive GHSOM to restrain the growing of hierarchy in GHSOM by reforming the map in each layer interactively. In order to verify the effectiveness of the proposed interface, we try to examine the Iris data set in \cite{UCI_IRIS}.

The remainder of this paper is organized as follows. In section \ref{sec:SOM}, we give a comprehension of basic SOM. In section \ref{sec:GHSOM}, the original algorithm of GHSOM is explained briefly. Section \ref{sec:unit-ge-eli} explains the unit generation/elimination process. Section \ref{sec:interactiveGHSOM} describes the algorithm of interactive GHSOM and its interface tool. Experimental results for classification of benchmark test in Section \ref{sec:experiments}. In Section \ref{sec:conclusion}, we give some discussions to conclude this paper.

\section{Self Organizing Map}
\label{sec:SOM}
\begin{figure}[htb]
\begin{center}
\includegraphics[scale=0.8]{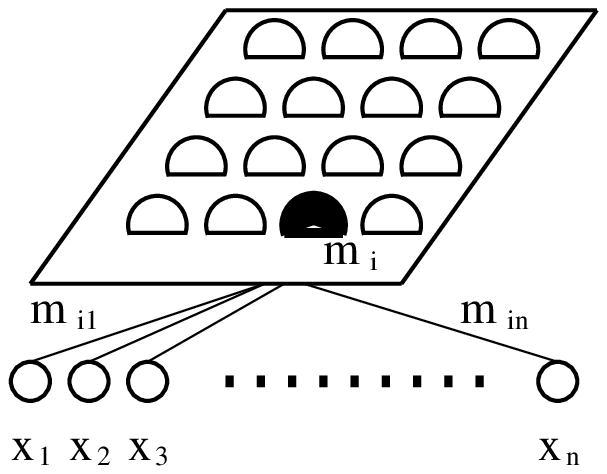}
\caption{An overview of SOM}
\label{fig:SOM}
\end{center}
\end{figure}

The basic SOM can be visualized as a sheet-like neural network array as shown in Fig.\ref{fig:SOM}. The cells (or nodes) of which become specifically tuned to various input signal patterns or classes of patterns in an orderly fashion. The learning process is competitive and unsupervised, which means that no teacher is required to define the correct output for an input. Only one map node called a winner node at a time is activated corresponding to each input. The map consists of a regular grid of processing units. A model of some multidimensional observations, eventually a vector consisting of features, is associated with each unit. The map attempts to represent all the available observations with optimal accuracy using a restricted set of models. At the same time the models become ordered on the grid so that similar models are close to each other and dissimilar models are far from each other.

A sequential regression process usually carries out fitting to the model vectors. The $n$ is the number of input signals. An input vector $\mbox{\boldmath $x$}$ is compared with all the model vectors $\mbox{\boldmath $m$}_{i}(t)$. The best-match unit on the map is identified. The unit is called the winner. For each sample $\mbox{\boldmath $x$}=\{x_{1}, x_{2}, \dots, x_{n}\}$, first the winner index $c$ (best match) is identified by the condition.

\begin{equation}
{}\parallel \mbox{\boldmath $x$}-{\mathbf m_{c}} \parallel = \min_i \parallel
\mbox{\boldmath $x$}- {\mathbf m_{i}} \parallel
\label{eq:som-compare}
\end{equation}

After that, all model vectors or a subset of them that belong to nodes centered 
around node $c$ are updated at time $t$ as

\begin{equation}
  \begin{array}{ll}
    {\mathbf m_{i}(t+1)} ={\mathbf m_{i}(t)}+h_{ci}(\mbox{\boldmath $x$}(t)-{\mathbf m_{i}(t)}) & \mathit{for}\:{}^{\forall}i \in N_{c}(t),\\
{\mathbf m_{i}(t+1)} = {\mathbf m_{i}(t)} &  \mathit{otherwise}
  \end{array}
\label{eq:som-modify}
\end{equation}

Here $h_{ci}()$ is the neighborhood function, a decreasing function of the distance between the $i$th and $c$th nodes on map grid. The $N_{c}(t)$ specifies the neighborhood around the winner in the map array. This regression is usually reiterated over the available samples. 

At the beginning of the learning process, the radius of the neighborhood is large and the range of radius becomes small according to the convergence state of learning. That is, as the radius gets smaller, the local correction of the model vectors in the map will be more specific. The $h_{ci}$  also decrease during learning.

\section{Growing Hierarchical SOM}
\label{sec:GHSOM}
\begin{figure}[tbp]
\begin{center}
\includegraphics[scale=0.6]{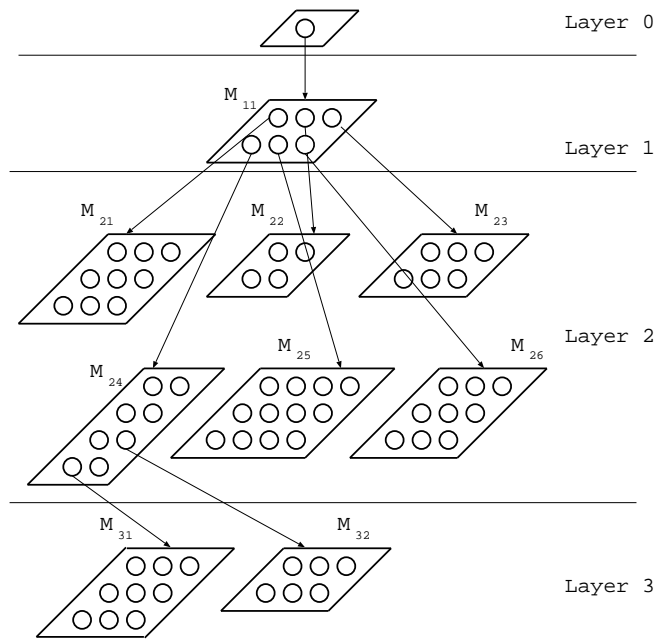}
\caption{A Hierarchy Structure in GHSOM}
\label{fig:overviewGHSOM}
\end{center}
\end{figure}

This section describes an basic algorithm of the growing hierarchical self organizing map (GHSOM)\cite{Rauber02}. The algorithm has been chosen for this application for its capability to develop a hierarchical structure of clustering and for the intuitive outputs which help the interpretation of the clusters. These capabilities allow different classification results from rough sketch to very detailed grain of knowledge. This technique is a development of the self organizing map (SOM), a popular unsupervised neural network model for the analysis of high dimensional input data \cite{Kohonen95}. Fig.\ref{fig:overviewGHSOM} shows the overview of hierarchy structure in GHSOM. Fig.\ref{fig:algorithmGHSOM} shows the algorithm of GHSOM.

\begin{figure}[tbp]
\begin{center}
\subfigure[generation patten a]{
\includegraphics[scale=0.7]{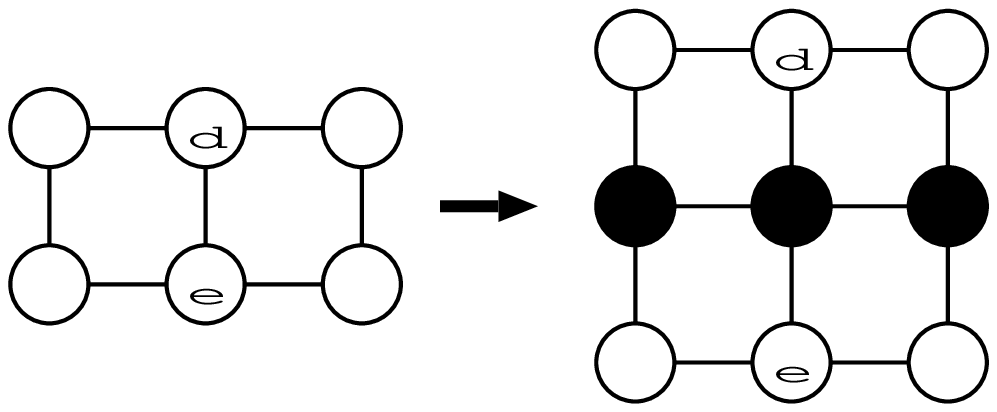}
}
\subfigure[generation patten b]{
\includegraphics[scale=0.7]{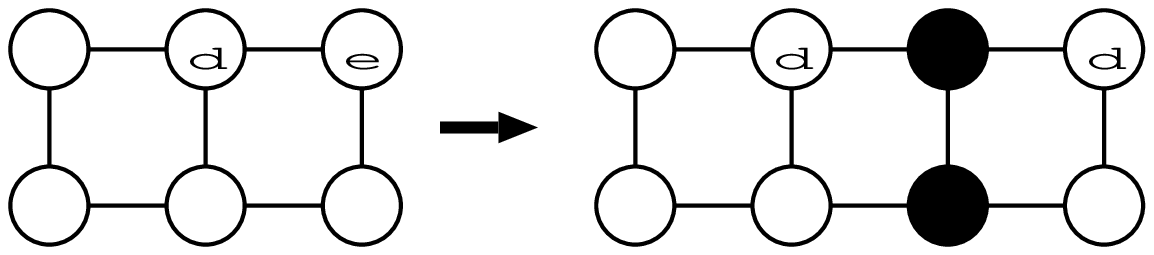}
}
\caption{Unit Generation Process in GHSOM}
\label{fig:generationGHSOM}
\end{center}
\end{figure}

\begin{center}
\begin{indentation}{0.1cm}{0.1cm}
\begin{breakbox}
\smallskip
\begin{enumerate}[Step 1)]
\item All parameters are initialized.
\item The mean quantization error of a unit $i$ is calculated in the $0$th layer in Fig.\ref{fig:overviewGHSOM}.
\begin{equation}
mqe_{0}=\frac{1}{n_{\it I}}\sum_{{\bf x}_{i} \in {\it I}}\parallel {\bf m}_{0} - {\bf x}_{i}\parallel, n_{\it I}= |{\bf{ \it I}}|
\label{eq:GHSOM-1}
\end{equation}
, where ${\bf m}_{0}$ is the average of input samples, $n_{\it I}$ is the number of input data set ${\it I}$, and ${\bf x}_{i}$ is an input vector for a sample.
\item Let ${\bf M}_{\ell,v}(v=1,2,\cdots)$ be a map in the $\ell(=1,2, \cdots)$ layer. The initial size of ${\bf M}_{\ell,v}$ is $u_{0}(=2\times2)$. The SOM clustering algorithm is employed in each map. The samples are divided into subcategories ${\it I}$ and let $k$ be an each winner unit in a subcategory.
\item Calculate the mean quantization error for each winner unit $k$ in ${\bf M}_{\ell,v}$ by using Eq.(\ref{eq:GHSOM-2}) and Eq.(\ref{eq:GHSOM-3}).
\begin{eqnarray}
\nonumber mqe_{k}=\frac{1}{n_{C}}\sum_{{\bf x}_{j} \in {\bf C_{k}}} \parallel {\bf m}_{k} - {\bf x}_{j}\parallel, \\
n_{\it C}= |{\bf C}_{k}|, {\bf C}_{k}\neq \phi
\label{eq:GHSOM-2}
\end{eqnarray}
\begin{equation}
qe_{k}=\sum_{{\bf x}_{j} \in {\bf C_{k}}} \parallel {\bf m}_{k} - {\bf x}_{j}\parallel, n_{\it C}= |{\bf C}_{k}|, {\bf C}_{k}\neq \phi
\label{eq:GHSOM-3}
\end{equation}
, where $mqe_{k}$ is the mean quantization error and $qe_{k}$ is the quantization error. The ${\bf m}_{k}$ is the reference vector of winner unit $k$, ${\bf C}_{k}$ is the samples which are allocated to the unit $k$, and ${\bf x}_{j}$ belongs to ${\bf C}_{k}$.
\item Let $e$ be a unit representing the largest error between input and reference vectors among winner units in ${\bf M}_{\ell,v}$. Compare the reference vector of the unit $e$ and its neighbor units, let $d$ be the unit with largest difference in the neighbor units.
\item Calculate the mean quantization error $mqe_{{\bf M}_{\ell,v}}$ for the subset ${\bf u}$ of a winner unit $k$ in ${\bf M}_{\ell,v}$ by using Eq.(\ref{eq:GHSOM-4})
\begin{equation}
mqe_{{\bf M}_{\ell,v}}=\frac{1}{n_{u}}\sum_{k \in {\bf u}} qe_{k}, n_{u}=\mid {\bf u}\mid
\label{eq:GHSOM-4}
\end{equation}
\item If the mean quantization error is satisfied with Eq.(\ref{eq:GHSOM-5}), units are inserted in row/column as shown in Fig.\ref{fig:generationGHSOM}.
\begin{equation}
mqe_{{\bf M}_{\ell,v}}\geq \tau_{1}qe_{w}
\label{eq:GHSOM-5}
\end{equation}
, where $w$ is the unit in the map of the $\ell -1$ layer.
\item The initial weights are given as the average of $d$ and $e$.
\item If the mean quantization error is not satisfied with Eq.(\ref{eq:GHSOM-5}), the insertion process of unit stops and returns to the map in the above layer and SOM is employed in the clustering in the map.
\label{item_insertion}
\item After Step \ref{item_insertion}), if the quantization error is larger(Eq.(\ref{eq:GHSOM-6})), then add a new layer to the map.
\begin{equation}
qe_{k}\geq \tau_{2}qe_{0}
\label{eq:GHSOM-6}
\end{equation}
\smallskip
\end{enumerate}
\end{breakbox}
\end{indentation}
\figcaption{The algorithm of GHSOM}
\label{fig:algorithmGHSOM}
\end{center}

\section{Unit Generation/Elimination}
\label{sec:unit-ge-eli}
The GHSOM adds new units in the array of row/column as shown in Fig.\ref{fig:generationGHSOM}. However, we observed some redundant units in the array to have no activation for any samples. Therefore, we propose a unit generation/elimination algorithm in the required position on the map.
For a normal NN, we can observe the following behaviors during learning phase.
\begin{description}
\item[Case1)] If an NN does not have enough units in order to satisfactorily infer the result, then the input weight vector will tend to fluctuate greatly even after a certain period of the learning process.
\item[Case2)] If an NN has enough units to infer and if the input weight vector of each unit converges to a certain value, it is possible to remove unnecessary units from the network without affecting the inference ability.
\end{description}

In the first case, the network needs to generate a new unit, as its parent's attribute is inherited. In the second case, it is necessary to delete a redundant unit in the calculation. Based on such behaviors, we can determine the conditions for the unit generation/elimination during the learning process.

\subsubsection{Unit generation}
The situation for unit generation appears when the representation capability of the network is insufficient. In general, given enough units in the hidden layer, an NN can form any mapping with any desired precision. Therefore, We can use the stabilized error as an index to determine whether the network needs to generate a new unit. If a stabilized error after a specified period is larger than the desired value, then a new unit may be generated.

The next problem is how to determine the position of the new unit in the hidden layer. The desirable position can be found through monitoring the behavior of the units in the hidden layer. If a unit strongly affects the final system error through the fluctuation in its input weight vector, then this unit requires a higher processing capability. If the input weight vector of a unit still fluctuates greatly even after an extended period of the learning process, this means that the unit does not have enough processing capability to learn the sub problem the unit must solve. Therefore, this unit can be split into two, i.e. we can add another unit to the exact same interconnection of the network, inheriting the original unit¡Çs attributes.

\begin{figure}[tbp]
\begin{center}
\includegraphics[scale=0.6]{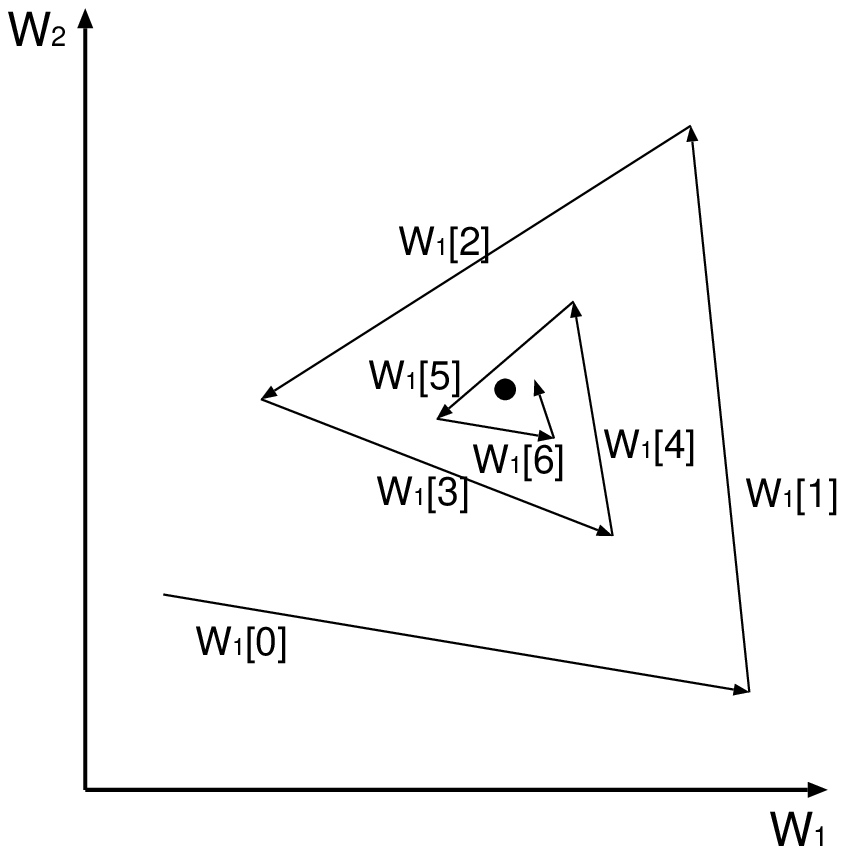}
\caption{Learning weights of a unit}
\label{fig:LearningWeight}
\end{center}
\end{figure}

Then we can give the following equation for generating a new unit. Fig.\ref{fig:LearningWeight} shows an idea of learning the weights of a unit by optimally searching the spaces of weights $\mathbf{W}_{1}$ and $\mathbf{W}_{2}$.

During the convergence process, the average length of movement between adjacent temporal weights gradually decreases. Based on this observation, we can define the following measure, called the walking distance($WD$) for units:
\begin{equation}
WD_{j}[n]=\sum_{m=1}^{n} \mathrm{Met}(\mathbf{W}_{j}[m], \mathbf{W}_{j}[m-1])
\label{eq:WD}
\end{equation}
,where $WD_{j}[n]$ is the $WD$ for unit $j$ in the hidden layer at iteration $n$, $\mathbf{W}_{j}[m]$ is the input weight to unit $j$ at an arbitrary iteration $m$, and $Met$ is some metric value that measures the distance between vectors in a metric space. In this subsection, $\mathrm{Met}()$ is given as an Euclidean metric:
\begin{eqnarray}
\mathrm{EuMet}(\mathbf{X}, \mathbf{Y}) = \sqrt[]{\mathstrut \sum_{i=1}^{N} (x_{i} -y_{i})^{2}}
\end{eqnarray}
where
\begin{equation}
N=\max(\mathrm{Dim}(\mathbf{X}), \mathrm{Dim}(\mathbf{Y}))
\end{equation}

In Eq.(\ref{eq:WD}), $WD_{j}$ is a measure for the time variance of the stored information for unit $j$, and can be considered the activity of unit $j$ in the parameter space. $WD$ that is too high for a particular unit indicates that the processing capability of the unit is insufficient, and a new unit is needed to share its load. Eq.(\ref{eq:WD}) can be approximated by the following operational measure:
\begin{eqnarray}
\nonumber WD_{j}[n]&=&\gamma WD_{j}[n-1]\\
\nonumber &+&(1-\gamma_{w})Met(\mathbf{W}_{j}[m], \mathbf{W}_{j}[m-1]),\\
&&\quad 0< \gamma_{w}<1
\end{eqnarray}

A unit $j$ should generate another unit if
\begin{equation}
\Delta \epsilon_{j} = \frac{\partial \epsilon}{\partial WD_{j}}\times WD_{j} > \theta_{G}
\end{equation}
, where $\epsilon$ is the overall system error, $WD_{j}$ is the walking distance for $j$, $\epsilon_{j}$ represents the contribution of unit $j$ to the overall system error, and $\theta_{G}$ is a threshold value.

\subsubsection{Unit elimination}
\label{sec:unitelimination}
In order to achieve a good overall system performance, we must remove unnecessary elements from the network. A network can gradually form an optimal structure in the development process by applying a mechanism with selection rules, which keeps effective units in the network and removes misplaced ones. To be more specific, we have based the selection mechanism on the following two observations:
\begin{description}
\item[(a)] If a unit does not form appropriate interconnections among other units, then it will become ineffective early in the development process.
\item[(b)] The units compete with each other for resources, because each unit tries to inhibit other units from taking exactly the same functional role that it plays in the network.
\end{description}

The above observations suggest the following guidelines for the unit elimination process. That is, we can remove a corresponding unit when:
\begin{description}
\item[(a)] It is not a functional element in the network.
\item[(b)] It is a redundant element of the network.
\end{description}

The criterion (a) can be checked by monitoring the output signal of a unit. As a measure for this criterion, we can use the variance of the output signal of a
unit $j$:
\begin{equation}
VA_{j}=(\bar{O}_{j}-O_{j})^{2}
\end{equation}
, where $O_{j}$ is the variance of the output signal level for unit $j$, and $\bar{O}_{j}$ is an average of $O_{j}$. We can define the operational measure of $VA_{j}$ as:
\begin{eqnarray}
\nonumber VA_{j}[n]=\gamma_{v}VA_{j}[n-1]+(1-\gamma_{v})(O_{j}-Act_{j}[n])^{2}, \\
\quad 0<\gamma_{v}<1
\end{eqnarray}
where
\begin{equation}
Act_{j}[n]=\gamma_{a}Act_{j}[n-1]+(1-\gamma_{a})O_{j}[n], \quad 0<\gamma_{a}<1
\end{equation}

$Act_{j}[n]$ is the operational measure of the average output signal for unit $j$.

If the $VA_{j}$ is zero for a given unit $j$, then this unit does not generate any additional information. In other words, this unit does not perform any signal processing function. Therefore, we can eliminate such a unit, when $VA_{j}$ is smaller than a certain threshold value in a practical sense.

The situation (b) can be identified by watching the correlation between the output values or units in the network. If two units are completely correlated, that is, given the output of one unit, the output of the other unit can be deduced with probability equal to 1, then the unit can be eliminated without affecting the performance of the network.

\subsubsection{Unit generation/elimination}
\begin{figure}[tbp]
\begin{center}
\subfigure[generation case 1]{
\includegraphics[scale=0.6]{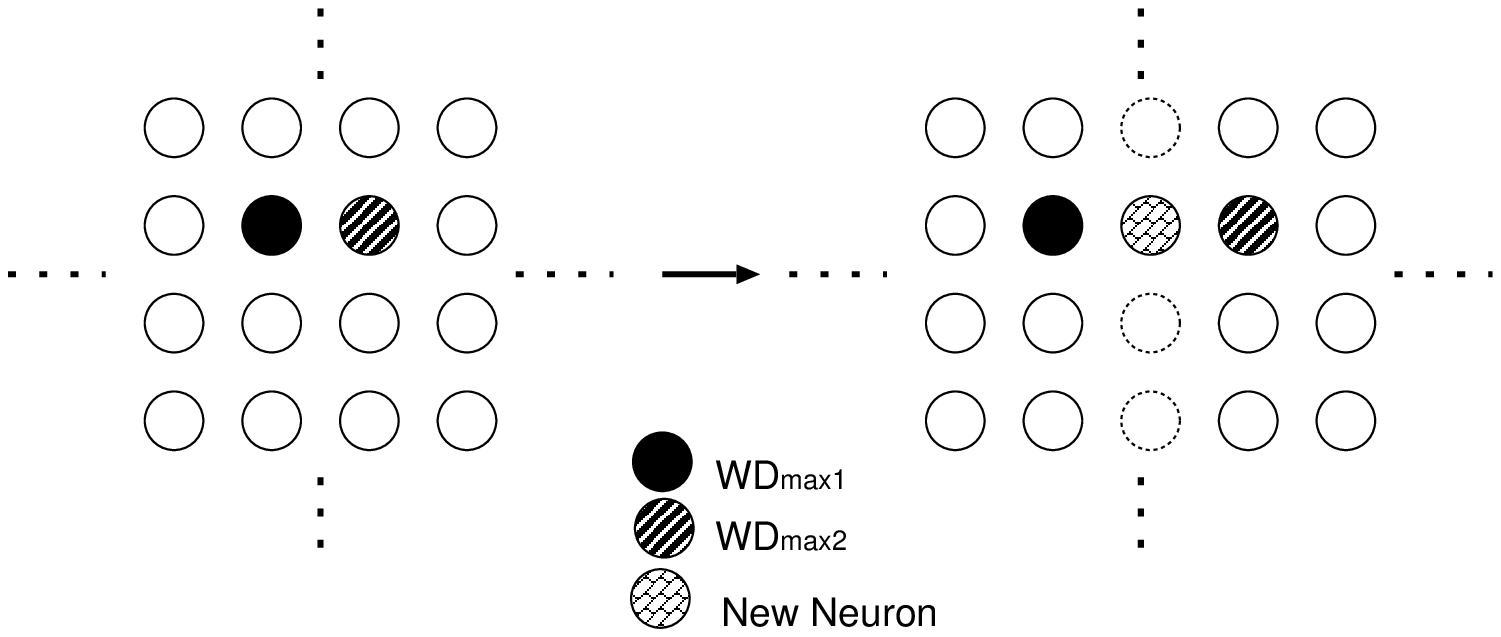}
\label{fig:LatticeUnitGeneration_a}
}
\subfigure[generation case 2]{
\includegraphics[scale=0.6]{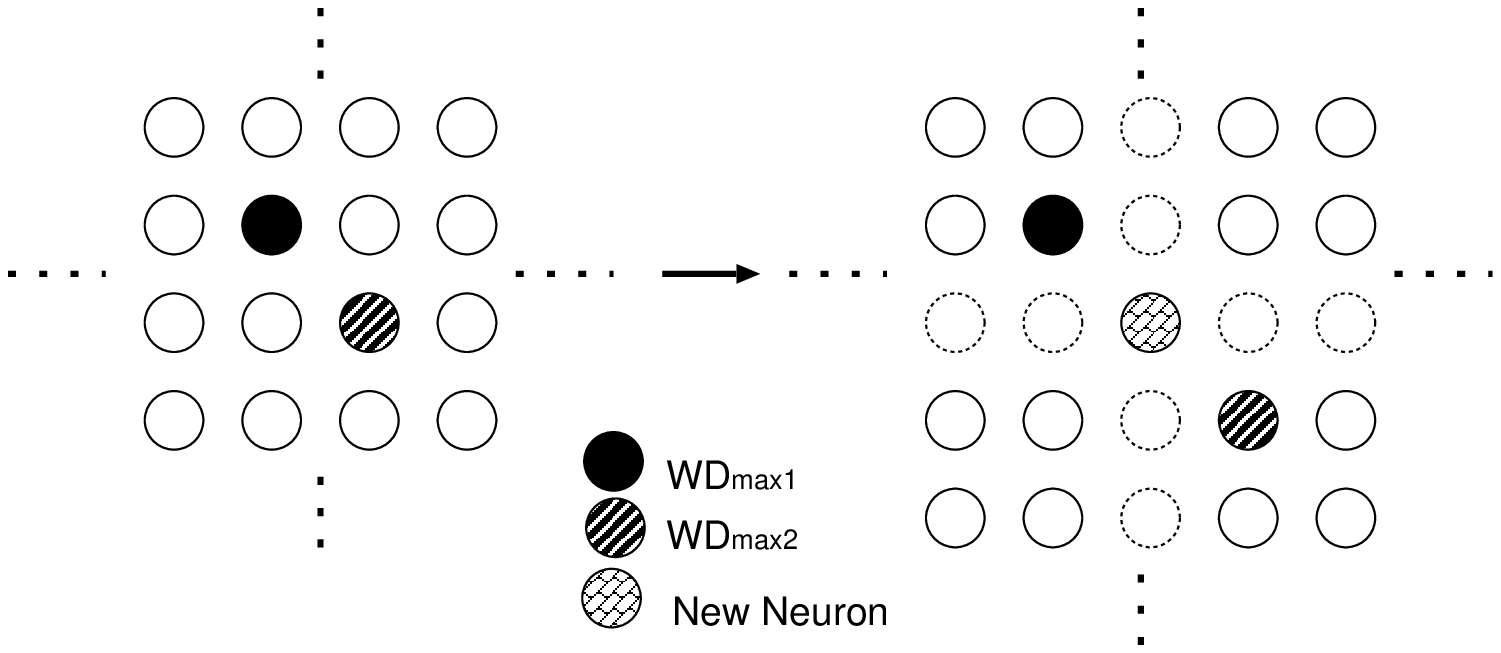}
\label{fig:LatticeUnitGeneration_b}
}
\caption{Unit Generation in map}
\label{fig:LatticeUnitGeneration}
\end{center}
\end{figure}

Fig.\ref{fig:LatticeUnitGeneration} shows implementation of unit generation process in the map. In the subsection, we define $WD_{max 1}$ as a unit with the largest $WD$ in the neighborhood $\mathcal{N}_{i}$, and $WD_{max 2}$ as a unit with the second largest $WD$. A new unit is added in the middle of two units $WD_{max 1}$ and $WD_{max 2}$. A new unit should set into the lattice according to the prior unit arrangement, because units are arranged in a plane.

\begin{figure}[tbp]
\begin{center}
\includegraphics[scale=0.6]{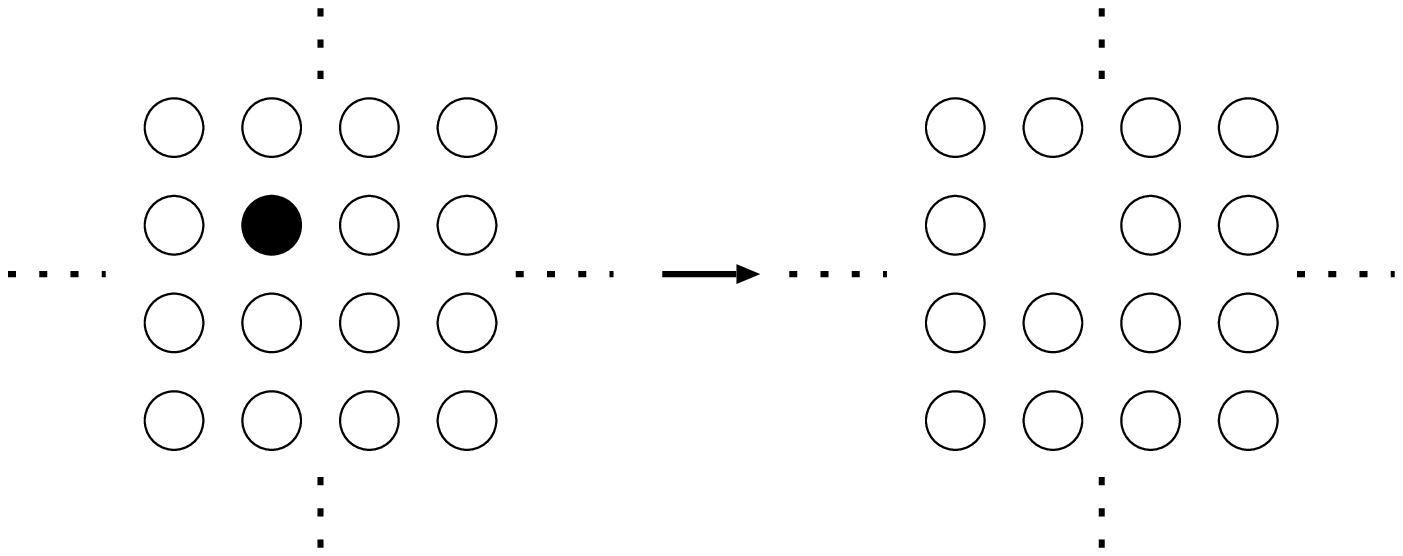}
\caption{Unit Elimination in map}
\label{fig:LatticeUnitElimination}
\end{center}
\end{figure}

The situation of unit generation is shown in Fig.\ref{fig:LatticeUnitGeneration}. Fig.\ref{fig:LatticeUnitGeneration_a} shows that the $WD_{max 1}$ unit is placed in a row or a column to the $WD_{max 2}$ unit. In this case, the new unit is added as shown on the right hand side of Fig.\ref{fig:LatticeUnitGeneration_a}. Fig.\ref{fig:LatticeUnitGeneration_b} shows that the $WD_{max 1}$ unit and the $WD_{max 2}$ unit are on a diagonal line. In this case, the new unit is added as shown on the right hand side of Fig.\ref{fig:LatticeUnitGeneration_b}. On the other hand, if the unit elimination rules are satisfied with the conditions described in subsection \ref{sec:unitelimination}, the specified unit is eliminated as shown in Fig.\ref{fig:LatticeUnitElimination}.

\section{Interactive GHSOM}
\label{sec:interactiveGHSOM}
The process of unit insertion and layer stratification in GHSOM works according to the value of $\tau_{1}$ and $\tau_{2}$. Because the threshold of their parameters gives a criteria to hierarchies in GHSOM, GHSOM cannot change its structure to data samples adaptively while training maps. Therefore, only a few samples are occurred in a terminal map of hierarchies. In such a case, GHSOM has a complex tree structure and many nodes(maps). As for these classification results, the acquired knowledge from the structure is lesser in scope or effect in the data mining. When we grasp the rough answer from the specification in the data set, the optimal set of parameters must be given to a traditional GSHOM. It is very difficult to find the optimal values through empirical studies.

We propose the reconstruction method of hierarchy of GHSOM even if the deeper GHSOM is performed. A stopping criteria for stratification is defined. Moreover, if the quantization error is large and the condition of hierarchies is not satisfied, the requirement for redistribution of error is defined.

\begin{description}
\item[Case 1)] If Eq.(\ref{eq:ichimuraGHSOM-1}) and the hierarchies are satisfied, stop the process of hierarchies and reinsert new units in the map.
\begin{equation}
n_{k}\leq \alpha n_{I}
\label{eq:ichimuraGHSOM-1}
\end{equation}
, where¡¤$n_{k}, n_{I}$ mean the number of input samples for the winner unit $k$ and of the all input samples${\it I}$, respectively. The $\alpha$ is a constant.
\item[Case2)] If Eq.(\ref{eq:GHSOM-6}) is not satisfied, the addition of layer is not generated, the quantization error in a unit is larger than the quantization error in a map, and Eq.(\ref{eq:GHSOM-6}) is satisfied, a new unit is inserted.
\begin{equation}
qe_{k}\geq \beta \tau_{1} \sum qe_{y}, y \in {\bf S}_{k}
\label{eq:ichimuraGHSOM-2}
\end{equation}
, where ${\bf S}_{k}$ is the set of winner units $k$.
\end{description}

\section{Experimental Results}
\label{sec:experiments}

\subsection{Iris data set}
The Iris data set\cite{UCI_IRIS} is the best known database to be found in the pattern recognition literature. The data set contains 3 kinds of classes with same ratio of class distribution(33.3\%); ``Iris Setosa'', ``Iris Versicolour'', and ``Iris Virginica''. One class is linearly separable from the other 2 classes. The latter are not linearly separable from each other. There are 4 kinds of input attributes; ``sepal length in cm'', ``sepal width in cm'', ``petal length in cm'', and ``petal width in cm''. The data set consists of 50 instances in each, where each class refers to a type of iris plant. Table \ref{table:ichimura-Irisdata} explains the summary of statistics for Iris data set.

\begin{table}[tbp]
\caption{Summary statistics for Iris data set}
\begin{tabular}{cccccc} \hline
class & Min & Max & Mean & SD & Class Correlation \\ \hline
sepal length & 4.3 & 7.9 & 5.84 & 0.83 & 0.7826 \\
sepal width  & 2.0 & 4.4 & 3.05 & 0.43 & -0.4194 \\
petal length & 1.0 & 6.9 & 3.76 & 1.76 & 0.9490 \\
petal width  & 0.1 & 2.5 & 1.20 & 0.76 & 0.9565 \\ \hline
\end{tabular}
\label{table:ichimura-Irisdata}
\end{table}

\subsection{Experimental Results}
\begin{figure}[tbp]
\begin{center}
\includegraphics[scale=0.45]{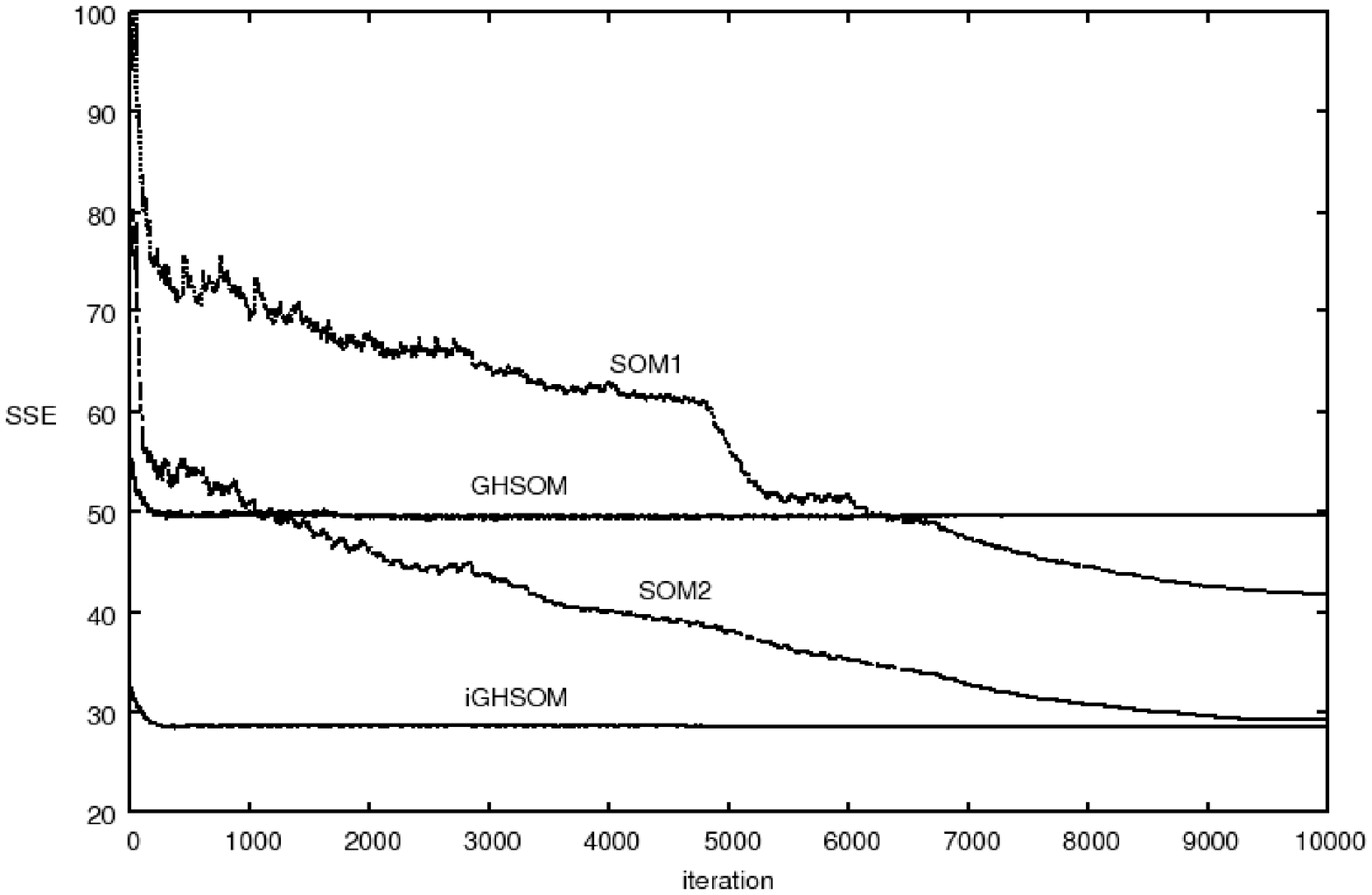}
\caption{Quantization Error for Iris data set}
\label{fig:error_GHSOM_iris0}
\end{center}
\end{figure}

Fig.\ref{fig:error_GHSOM_iris0} shows the quantization error in SOM, traditional GHSOM, the proposed GHSOM, respectively. The map size of 'SOM1' and 'SOM2' as shown in Fig.\ref{fig:error_GHSOM_iris0} were $12\times 10$, $20\times20$, respectively. In this experiment, $\tau_{1}$ was 0.07 and $\tau_{2}$ was 0.01, $\alpha$ was 0.04, $\beta$ was 4. The computation time of proposed method was 11.49 seconds, although that of SOM2 was 11.49 seconds, 18.50 seconds. Table \ref{table:AIC-Irisdata} shows the value of AIC by using the method described in \cite{Moreno2003}.

\begin{table}[tbp]
\caption{An Information Criteria for Interactive GHSOM}
\begin{tabular}{|c|c|c|c|} \hline
\multicolumn{2}{|c|}{Quantization Error} & \multicolumn{2}{c|}{AIC} \\ \hline
GHSOM & Interactive GHSOM & GHSOM & Interactive GHSOM \\ \hline
1.012 & ${\bf 0.836}$ & 178.97 & ${\bf 114.81}$ \\ \hline
 \end{tabular}
\label{table:AIC-Irisdata}
\end{table}

\subsection{Interface of Interactive GHSOM}
\begin{figure}[tbp]
\begin{center}
\includegraphics[scale=0.45]{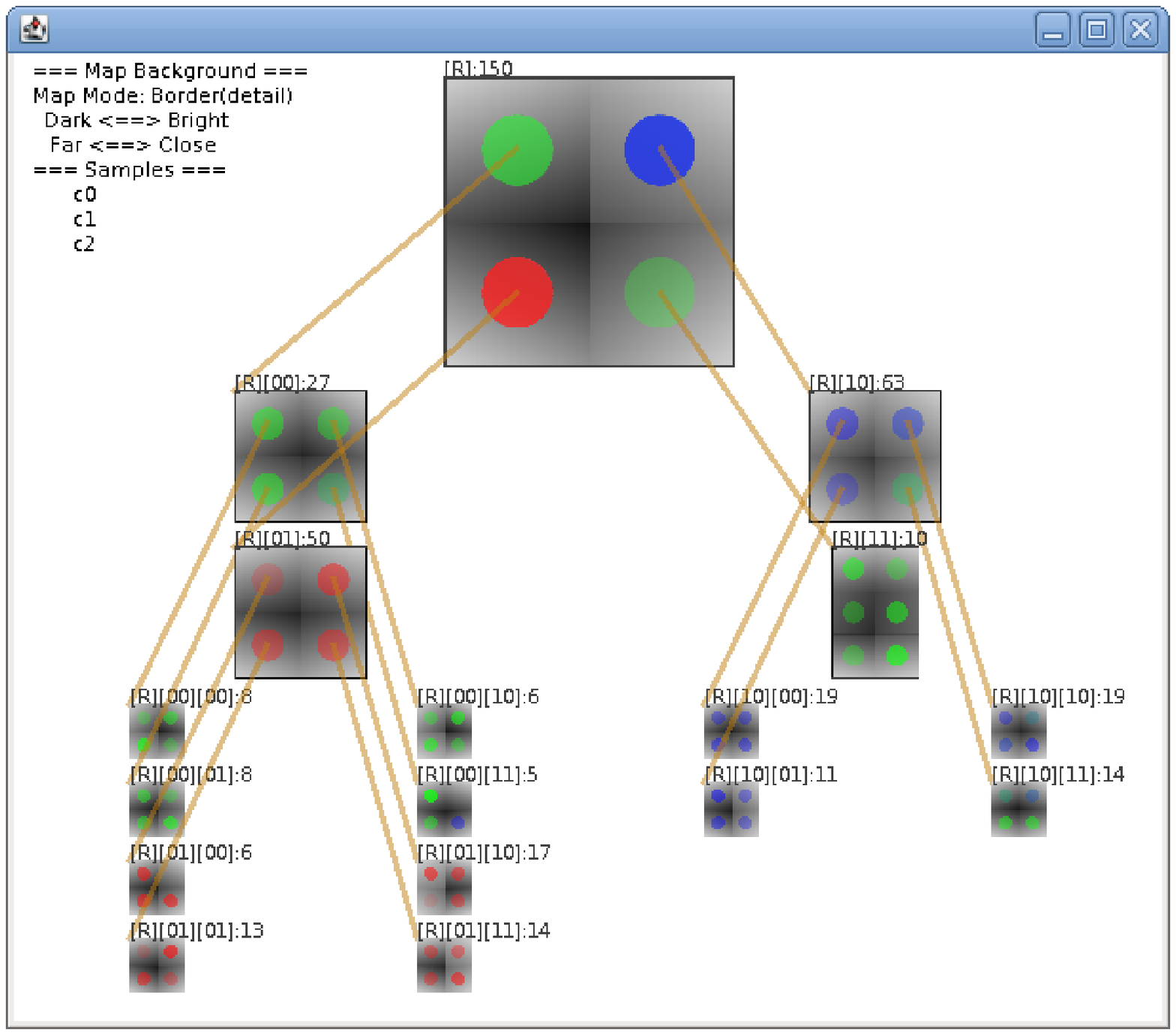}
\caption{Simulation Result for Iris data set}
\label{fig:GHSOM_iris0}
\end{center}
\end{figure}

\begin{figure}[tbp]
\begin{center}
\includegraphics[scale=0.6]{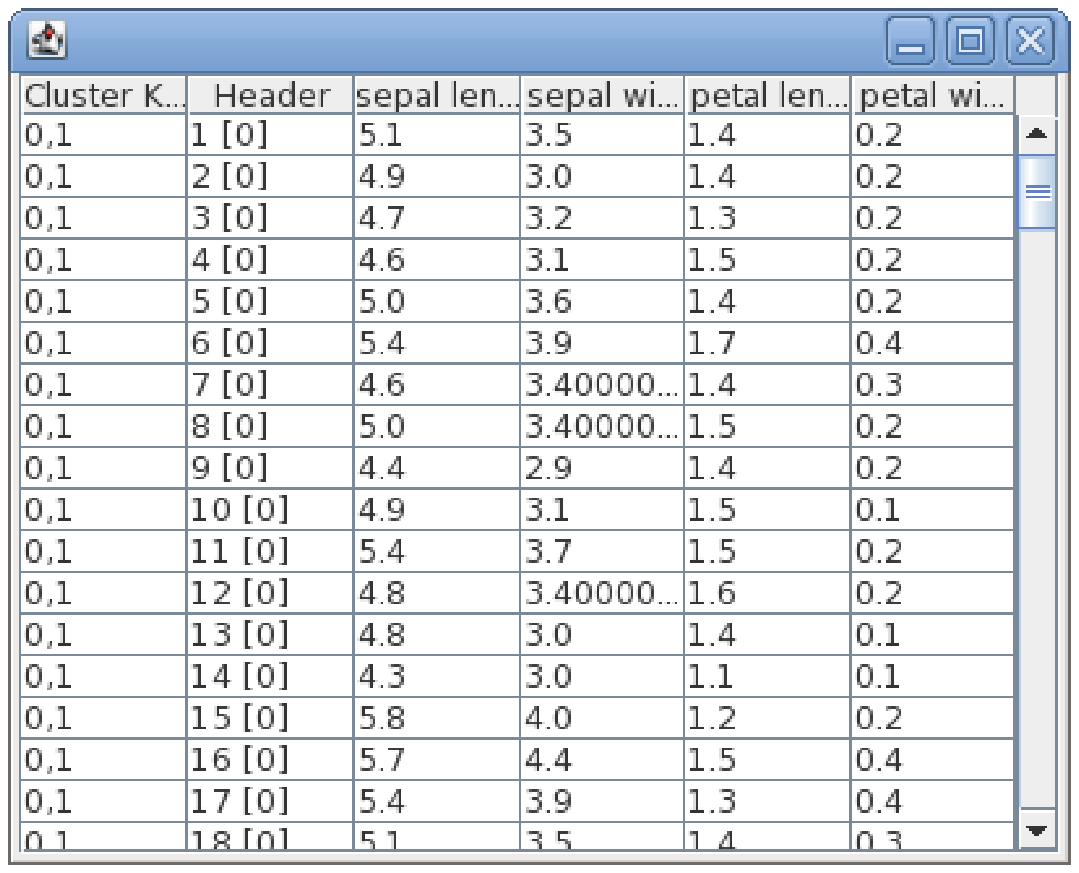}
\caption{A table for clustering result}
\label{fig:GHSOM_iris_table}
\end{center}
\end{figure}

We developed the interface of Interactive GHSOM to acquire the knowledge intuitively. This tool was developed by Java language. Fig. \ref{fig:GHSOM_iris0} shows the clustering results by GHSOM. When we touch the unit in a map, the other window as shown in Fig. \ref{fig:GHSOM_iris_table} are displayed and the samples in allocated in the corresponding unit listed in the table. The color of unit shows the pattern of sample in RGB. The similar color of unit means the similar pattern of samples. If the number of units in a map are increased, a few of samples cannot be classified into. Once the unit connected to the map is selected, the method re-calculates to find an optimal set of weights in the local tree structure search and then a better structure is depicted. 

\begin{figure}[tbp]
\begin{center}
\subfigure[Case 1]{
\includegraphics[scale=0.45]{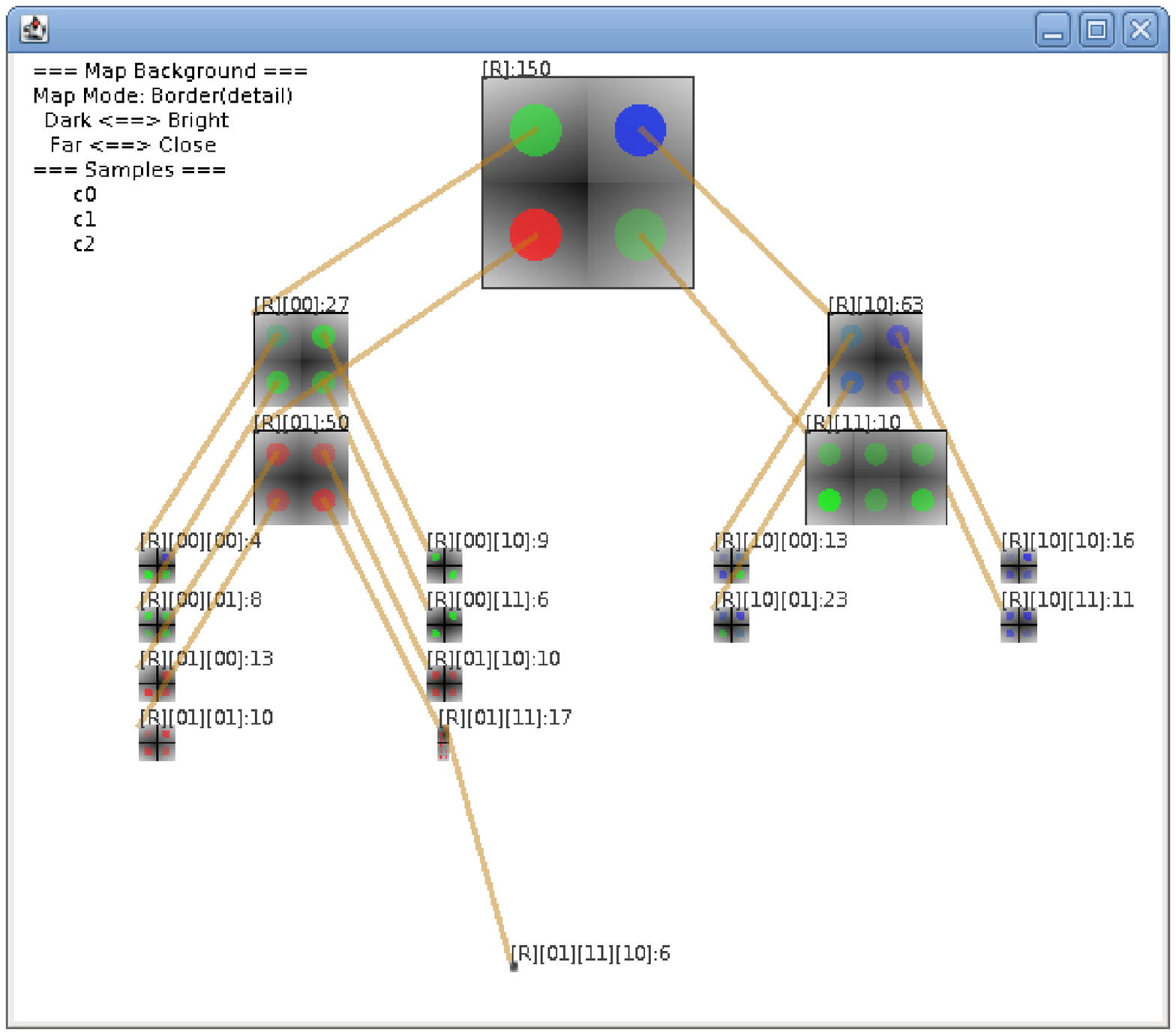}
\label{fig:GHSOM_iris1}
}
\subfigure[Case 2]{
\includegraphics[scale=0.45]{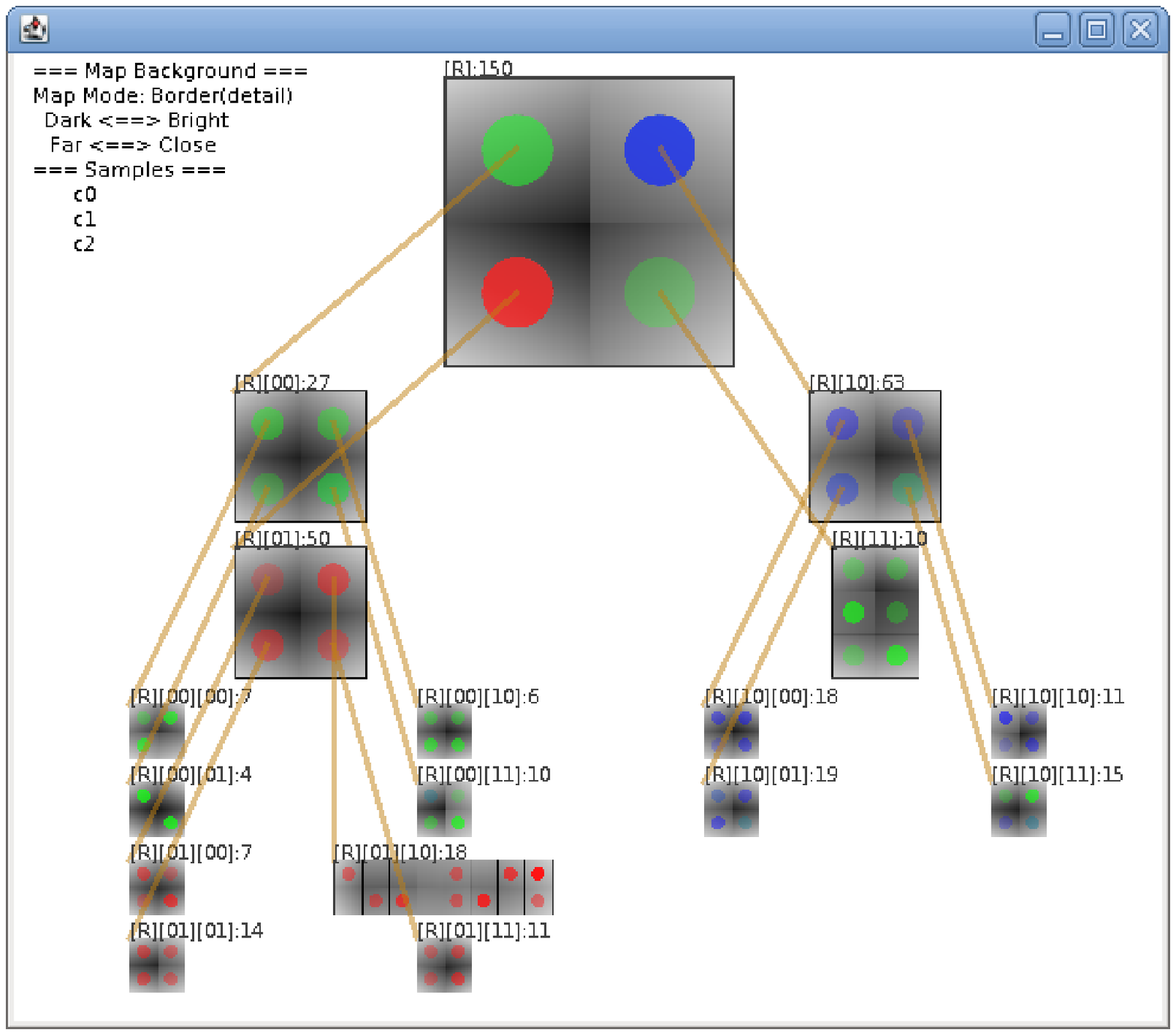}
\label{fig:GHSOM_iris2}
}
\caption{Simulation Result by Interactive GHSOM}
\label{fig:GHSOM_iris_after}
\end{center}
\end{figure}

Figs. \ref{fig:GHSOM_iris1} and \ref{fig:GHSOM_iris2} show the re-clustering results by using interactive GHSOM. The left side tree in Fig.\ref{fig:GHSOM_iris1} is developed and more detailed knowledge description was appeared. On the other hand, the left side tree in Fig.\ref{fig:GHSOM_iris2} was classified in the same map.

\section{Conclusive Discussion}
\label{sec:conclusion}
In this paper, we proposed the clustering method of interactive GHSOM that the layer stratification in GHSOM is controlled and the unit insertion in its original layer. The threshold for 2 kinds of parameters in traditional GHSOM must be defined and the values cannot change adaptively and the structures may be obtained by predefined parameters. However, when we consider the detailed substructure focusing the tree structure, the calculation should be started in the initial process with parameter settings. In order to improve such problems, the innovative process between computer simulation and human consideration is required and the interactive GHSOM is an effective clustering method to discover the concurrent knowledge acquisition. The empirical tests to practical data sample such as marketing, fraud detection and medical information has been studied in near future.

\end{document}